\documentclass{article} 
\usepackage{iclr2027_conference,times}

\usepackage[colorlinks=true,citecolor=blue,linkcolor=blue]{hyperref}
\usepackage{graphicx}
\usepackage{tabularx}
\usepackage{booktabs}
\usepackage{wrapfig}
\usepackage[nohyperlinks, nolist]{acronym}

\usepackage{amsmath,amsfonts,bm}

\def\Figref#1{Figure~\ref{#1}}

\def\eqref#1{equation~\ref{#1}}
\def\Eqref#1{Equation~\ref{#1}}

\def\1{\bm{1}}

\def\rd{{\textnormal{d}}}

\def\rva{{\mathbf{a}}}

\def\rvs{{\mathbf{s}}}

\def\vtheta{{\bm{\theta}}}

\def\vphi{{\bm{\phi}}}

\def\va{{\bm{a}}}

\def\vs{{\bm{s}}}

\DeclareMathAlphabet{\mathsfit}{\encodingdefault}{\sfdefault}{m}{sl}
\SetMathAlphabet{\mathsfit}{bold}{\encodingdefault}{\sfdefault}{bx}{n}

\def\gA{{\mathcal{A}}}
\def\gB{{\mathcal{B}}}

\def\gD{{\mathcal{D}}}

\def\gJ{{\mathcal{J}}}

\def\gP{{\mathcal{P}}}

\def\gS{{\mathcal{S}}}

\def\sD{{\mathbb{D}}}

\def\sR{{\mathbb{R}}}

\newcommand{\E}{\mathbb{E}}

\newcommand{\KL}{\sD_{\mathrm{KL}}}

\DeclareMathOperator*{\argmax}{arg\,max}
\DeclareMathOperator*{\argmin}{arg\,min}

\begin{acronym}[Notation] 
\acrodef{VLA}[VLA]{vision-language-action model}
\acrodef{PPO}[PPO]{proximal policy optimization}
\acrodef{IRL}[IRL]{inverse reinforcement learning}
\acrodef{SAC}[SAC]{soft actor-critic}
\acrodef{BC}[BC]{behavioural cloning}
\acrodef{LBM}[LBM]{large behaviour model}
\acrodefplural{LBM}[LBMs]{large behaviour models}
\acrodef{CSIL}[CSIL]{coherent soft imitation learning}
\acrodef{RL}[RL]{reinforcement learning}
\end{acronym}

\acused{BC}

\usepackage{amsthm}
\newtheorem{theorem}{Theorem}

\newtheorem{definition}{Definition}

\usepackage[ruled]{algorithm2e}

\SetCommentSty{mycommfont}

\usepackage{tikz}
\newcommand{\cyanx}{%
  \tikz[baseline=-0.5ex]\draw[cyan]
    plot[mark=x,mark size=3pt,mark options={line width=1pt}] (0,0);%
}

\title{Coherent Off-Policy Improvement of\\
Large Behavior Models with\\Learned Rewards
}

\author{Christian Scherer$^{1,3}$, Joe Watson$^2$, Theo Gruner$^{1,4}$, \\
\textbf{Daniel Palenicek$^{1,4}$,}
\textbf{Ingmar Posner$^2$,} 
\textbf{Jan Peters$^{1,4,5,6}$}\\
  $^1$Technical University of Darmstadt,
  $^2$University of Oxford,\\
  $^3$Zuse School ELIZA,
  $^4$hessian.AI,
  $^5$German Research Center for AI (DFKI),\\
  $^6$Robotics Institute Germany (RIG),\\
  \texttt{joe@robot-learning.de}
  \vspace{-2em}
}

\iclrfinalcopy 
\begin{document}

\maketitle

\begin{abstract}
Distilling expert demonstration data into large generative models using behavioral cloning is a scalable approach to learning capable policies for robotic control, particularly for dexterous manipulation.
Reinforcement learning (RL) can be used as a means to finetune these policies further using additional experience. 
An open question is whether RL is more sample-efficient than collecting more expert demonstrations.
Prior work has finetuned large pretrained policies in a scalable fashion by applying RL to a smaller residual policy that corrects the pretrained model. 
However, for typical sparse reward tasks, RL algorithms can struggle to optimize the behavior in a sample-efficient manner.
We explore inverse reinforcement learning, where a dense reward function is learned from expert demonstrations, potentially reducing the challenge of RL finetuning.
We specifically consider coherent imitation learning, an IRL method that facilitates improvement of the BC policy by using a specific learned reward formulation with theoretical guarantees for imitation learning.   
We show that our IRL method maintains or improves the performance of \texttt{pi-0.5} on all six sparse manipulation tasks and achieves a $\geq 90\%$ success rate on five out of six complex manipulation tasks, outperforming RL-based baselines using sparse rewards.
We also demonstrate real-world learning on a robot manipulation task, finetuning \texttt{pi-0.5} to success with less than 30 minutes of environment interaction time.
By ensuring our initial pretrained finetuning policy is optimal for our initial reward and critic, our method circumvents the initial drop commonly seen in RL finetuning and enables faster improvement.
\end{abstract}


\section{Introduction}
Distilling massive expert demonstration datasets into large behavior models (LBMs) using behavioral cloning (\ac{BC}) has seen large success in robotic control, yielding highly capable policies for complex, long-horizon manipulation \citep{driess2023palm, barreiros2025careful} given enough high-quality demonstration data.
Despite these advances, covariate shift due to accumulating errors remains a fundamental limitation of any \ac{BC} policy.
Solutions such as collecting supplementary recovery demonstrations are fundamentally limited because they rely on teleoperators to anticipate out-of-distribution states.
Since online agents will inevitably face novel states, online adaptation becomes indispensable.

\begin{wrapfigure}{r}{0.42\textwidth}
      \vspace{-0.5em}
      \begin{center}
         \includegraphics[width=0.4\textwidth]{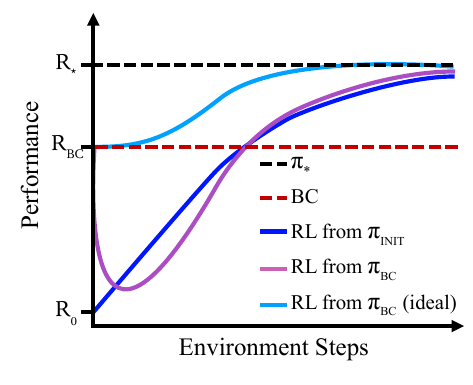}
      \end{center}
    \caption{An illustration of RL finetuning of strong BC policies using residuals. 
    Despite a strong initial performance by a BC policy, using function approximation in actor-critic methods means this performance is rapidly unlearned and relearned, so in practice, the BC initialization provides little benefit when running the RL method from scratch. 
    We use coherent soft imitation learning (CSIL) for (inverse) RL finetuning without unlearning due to the learned coherent reward.
    }
    \label{fig:hero}
    \vspace{-3em}
\end{wrapfigure}

\Ac{RL} \citep{sutton2018reinforcement} is a natural candidate to improve \ac{BC} policies through online experience. In particular, residual \ac{RL} offers an attractive set of characteristics for online adaptation: freezing the base policy and learning a correction allows for parameter-efficient finetuning of even large models.
Recent efforts to improve online adaptation have introduced fine-grained corrections for action chunks \citep{ankile2025imitation}, structured exploration \citep{yuan2025decorator}, and more efficient value learning \citep{ma2026makesvaluelearningefficient}. Despite these advances, two major challenges remain for \ac{RL} for foundation models: First, designing dense, task-specific rewards for diverse, open-world tasks is difficult and prone to misspecification \citep{eschmann2021reward}.
Second, even in parameter-efficient paradigms like residual \ac{RL} \citep{xiao2026_pld}, where the base model is frozen, the framework remains vulnerable to initial performance degradation, as illustrated in \Figref{fig:hero} and shown experimentally in Figure \ref{fig:lbm-results}.
Crucially, this unlearning directly degrades the sample efficiency of any RL finetuning. 

To achieve online adaptation without degrading the foundation model (see \Figref{fig:hero}), we build on CSIL \citep{watson2023coherent}, an entropy-regularized IRL algorithm.
In contrast to other (deep) IRL algorithms, CSIL is designed to finetune a pretrained behavior policy using IRL due to its specially designed reward structure, as IRL methods with black-box reward models tend to unlearn the BC behavior rather than improve it \citep{watson2023coherent}. 
Our contributions are:
\begin{enumerate}
    \item We propose using inverse RL, specifically CSIL, to finetune LBMs with a learned dense reward as an alternative to RL with challenging sparse rewards.
    \item We provide an improved implementation of CSIL based on recent advances in deep actor-critic methods using batch normalization, weight normalization, and categorical critics \citep{palenicek2025xqc}.
    \item We present an experimental benchmark using \texttt{pi-0.5} across six sparse, simulated manipulation tasks, showing benefits over RL-based finetuning baselines.
    We also demonstrate successful real-world learning on a robot manipulation task.
\end{enumerate}

\section{Related Work}
\label{sec:related}
While LBMs show strong performance, we seek to further improve them in an autonomous fashion using RL.
The challenge here is that LBMs have incredibly large parameter spaces, which are tractable for supervised learning but are a challenge for reinforcement learning, where the gradient updates have higher variance.

\citet{ren2025diffusion} refine a diffusion policy using on-policy RL by augmenting environment rollouts with those generated by the diffusion model, which allows them to optimize a subset of denoising steps rather than the entire model. 
\citet{chen2026pitextttrlonlinerlfinetuning} propose a similar approach, but adapted for flow-matching policies.
\citet{ankile2025imitation} also use on-policy RL to improve a diffusion policy using a residual MLP policy focusing on precise assembly and sim-to-real transfer.
On-policy methods are challenging to apply to real-world experience due to their poor sample efficiency.
As a result, prior work has investigated off-policy \ac{RL}.
\citet{wagenmaker2025steering} propose diffusion steering via RL, which applies off-policy RL to the latent space of the diffusion model rather than the model itself.  
For general LBMs, residual RL \citep{silver2018residual} is attractive to finetune an additional, smaller policy. 
\citet{xiao2026_pld} propose `probe, learn, distill' (PLD), which combines regularized offline RL critics for off-policy RL with additional probing and distillation steps.
\citet{ankile2025residual} propose ResFit, which also uses off-policy RL to finetune diffusion policies in a residual fashion.
While these results validate the potential of residual RL finetuning, the presented learning curves consistently report severe initial performance drops during early finetuning.
This phenomenon is due to the learning dynamics of actor-critic methods, where gradient updates from the initial suboptimal critic `unlearn' the BC behavior rather than finetune it (\Figref{fig:hero}).
BC unlearning has also been reported in the inverse RL setting due to the jointly learned reward function \citep{orsini2021matters,watson2023coherent}. 
%
%
%

In concurrent work, \citet{sun2026prior} also apply off-policy RL to finetune residual policies, but use a heuristic and adaptive BC-based policy regularization using the learned $Q$-values.
While their method, DICE-RL, does not exhibit unlearning due to the policy regularization, our proposed method appears x3 more sample-efficient on image-based Robomimic tasks.
\citet{zhong2026vla} improve LBMs using on-policy distillation from a provided expert policy. 
This approach strongly resembles CSIL, as the reward is a log-policy ratio, though the on-policy formulation is different.


\section{Coherent Inverse Reinforcement Learning}
\label{sec:csil}

In this section, we summarize the coherent soft imitation learning algorithm and describe several adaptations that scale the algorithm to the harder tasks where LBMs benefit from finetuning.

\subsection{Coherent Soft Imitation Learning}
We consider a Markov decision process (MDP) $\langle\gS, \gA, \gP, r, \gamma, \mu_0\rangle$, where $\gS$ is the state space, $\gA$ is the action space, $\gP{\;:\;}\gS{\,\times\,}\gA{\,\times\,}\gS{\,\rightarrow\,}\sR^+$ is the transition model, $r{\;:\;}\gS{\,\times\,}\gA{\,\rightarrow\,}\sR$ is the reward function, $\gamma$ is the discount factor, and $\mu_0$ is the initial state distribution
${\vs_0\sim\mu_0(\cdot)}$.
When the reward is unknown, imitation can be performed using a dataset $\gD$ of transitions $(\vs,\va,\vs')$ as demonstrations.
The policy could be inferred directly or by inferring a reward and policy jointly, referred to as behavioral cloning and inverse reinforcement learning, respectively. 

We consider the KL-regularized RL setting, where policy $\pi$ is optimized to maximize the state-action value function $Q$ subject to a soft KL constraint against a prior policy $p(\va|\vs)$ to promote exploration,
$\max_\pi \E_{\vs,\va \sim \rho_q}[Q(\vs,\va)] - \alpha\,\KL[
\pi(\va|\vs)\mid\mid p(\va|\vs)]$.
The optimal policy update blends the exponentiated advantage function `pseudo-likelihood' with the prior, as a form of regularized Boltzmann policy \citep{haarnoja2018softactorcriticoffpolicymaximum} with temperature $\alpha > 0$ and soft value function $V(\vs)$,
\begin{align}
\pi(\va|\vs)
\propto
\exp((Q(\vs,\va){\,-\,}V(\vs))/\alpha
)\,p(\va|\vs),
\;
V(\vs) = \alpha\log\textstyle\int \exp(Q(\vs,\va)/\alpha)\,p(\va|\vs)\,\rd\va.
\label{eq:posteriorpolicy}
\end{align}
In the function approximation setting, the update is performed in a variational fashion by minimizing the reverse KL divergence between the parametric policy $\pi_\vtheta$ and the critic-derived update at sampled states $\vs$, as done in the soft actor-critic (SAC) algorithm \citep{haarnoja2018softactorcriticoffpolicymaximum},
\begin{align}
\vtheta_*
&= 
\textstyle\argmin_\vtheta
\textstyle
\E_{\vs\sim\gB}[\sD_{\textsc{kl}}[
\pi_\vtheta(\cdot\mid\vs)
\mid\mid
\exp((Q(\vs,\cdot) - V(\vs))/\alpha)
\,p(\cdot|\vs)]],\nonumber\\
&=
\textstyle
\argmax_\vtheta \gJ_\pi(\vtheta),
\,
\gJ_\pi(\vtheta)
=
\E_{
\vs\sim \gB,\,
\va \sim \pi_\vtheta(\cdot|\vs)
}
\left[
Q(\vs,\va)
- \alpha\,
(\log \pi_\vtheta(\va|\vs)
-
\log
p(\va|\vs))
\right].
\label{eq:sac-objective}
\end{align}

\begin{figure}[!b]
    \vspace{-1em}
    \centering
    \begin{minipage}{0.51\textwidth}
    \begin{definition}
        (Coherent reward,~\citet{watson2023coherent}).
        For a BC policy $\pi$, prior $p(\va\mid\vs)$ and temperature $\alpha > 0$, the scaled log-policy ratio
        \begin{align}
            \tilde r(\vs,\va) = \alpha \left(\log \pi(\va\mid\vs)
            - \log p(\va \mid \vs )\right),
        \end{align}
        is the coherent reward function, where $\pi$ is the policy and $p$ is a uniform prior.
    \label{def:csil}
    \end{definition}
    \begin{theorem}{(Coherent reward, \citet{watson2023coherent})}
    In the KL-regularized Bellman equation, 
    the log-policy ratio (Definition \ref{def:csil}) is derived as
    \begin{align*}
        \alpha \log \frac{\pi(\va\mid\vs)}{p(\va\mid\vs)} = r(\vs,\va) + \gamma\E[V(\vs')] - V(\vs),
    \end{align*}
    which shows it is a shaped reward (Theorem \ref{th:ng}).
    \label{th:csil}
    \end{theorem}
    
    \begin{theorem}{(Reward shaping, \citet{ng99})} For potential $\psi$, the shaped reward $\tilde r(\vs,\va)$, where
    \begin{align*}
        \tilde r(\vs,\va) = r(\vs,\va) + \gamma\E[\psi(\vs')] - \psi(\vs)
    \end{align*}
    has the same optimal policy as $r(\vs,\va)$.
    \label{th:ng}
    \end{theorem}
    \end{minipage}
    \hfill
    \begin{minipage}{0.45\textwidth}
    \includegraphics[width=\linewidth]{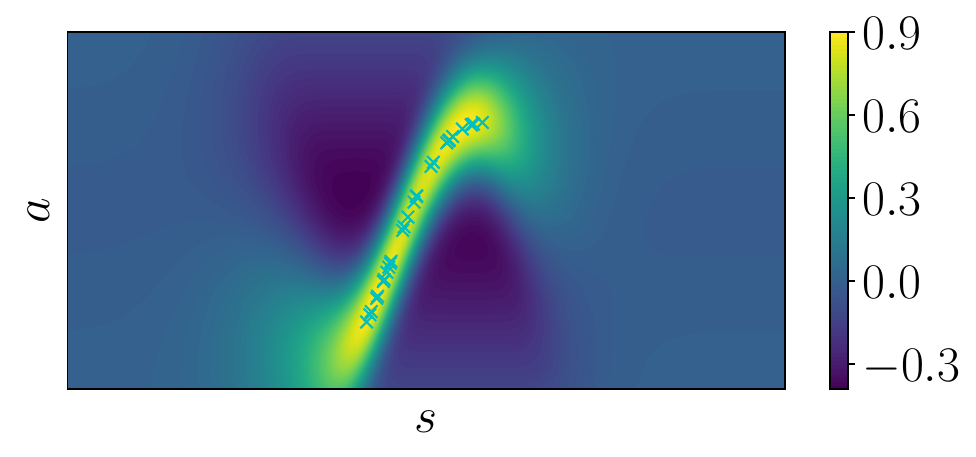}
    \caption{Given demonstration data (\protect\cyanx), the coherent reward provides positive rewards for correct actions for observed states, negative rewards for incorrect actions in seen states, and zero rewards for any action under unseen states.
    This reward encourages the agent to stay (and return) to the demonstration distribution.
    In contrast to adversarial methods, no on-policy samples are needed to learn it.
    The contour plot is taken from \citet{watson2023coherent}.
    }
    \label{fig:csil_reward}
    \end{minipage}
\end{figure}
The above objective $\gJ_\pi$ can be maximized using reparameterized gradients and minibatches from the replay buffer $\gB$ \citep{haarnoja2018softactorcriticoffpolicymaximum}.
In SAC, $p(\va|\vs)$ is a uniform distribution to provide maximum entropy regularization.

In coherent soft imitation learning (CSIL) \citep{watson2023coherent}, inverse RL is achieved by rearranging the policy update in \Eqref{eq:posteriorpolicy} and substituting it into the KL-regularized Bellman equation. 
This policy update inversion results in the log-policy ratio appearing as a reward term in the Bellman equation.
By substituting a BC policy as the optimal policy, the log-policy ratio becomes a learned reward that is \textit{coherent} with the BC policy $\pi$ (Definition \ref{def:csil}).
Theorem \ref{th:csil} shows that this coherent reward is a shaped variant of the true reward and therefore shares the same optimal policy (Theorem \ref{th:ng}, \citet{ng99}).
The benefit of this approach to IRL is that BC policies are naturally finetuned with additional experience, in contrast to methods where the reward and policy are learned jointly, and BC behavior is quickly unlearned \citep{watson2023coherent}.
The log-policy ratio can also be interpreted as a dense reward that encourages the agent to follow (and return to) the demonstration distribution (\Figref{fig:csil_reward}).

CSIL requires a policy architecture that is \textit{stationary}, such that its predictions revert to its prior $p(\va\mid\vs)$ outside the training distribution, ensuring the coherent reward provides a neutral reward of zero in unseen states. Following \citet{watson2023coherent}, we use the HetStat (heteroscedastic stationary) architecture, which achieves stationarity via a periodic activation function in a wide final layer \citep{meronen2021periodicactivationfunctionsinduce}, approximating a stationary kernel via the Wiener-Khinchin theorem \citep{gpml}.
Moreover, while CSIL is an actor-critic method that builds on SAC, it regularizes the policy against the BC policy rather than providing maximum entropy regularization.
This KL regularization has a temperature $\beta$ which does not have to match $\alpha$. The latter is typically set to make the coherent reward scale invariant \citep{watson2023coherent}.  

\begin{figure}[t]
    \centering
    \includegraphics[width=\linewidth]{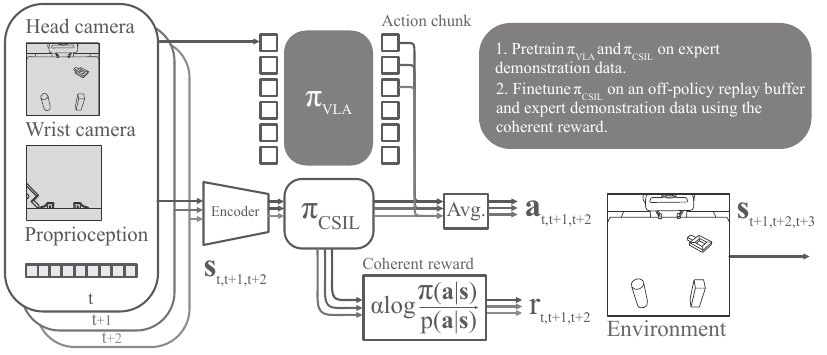}
    \caption{A system-level figure of CSIL-based finetuning of VLAs using averaged actions. The CSIL policy reacts to every observation and adapts the action chunk of the VLA, which executes at the action chunk frequency. CSIL has a dedicated encoder and does not use the VLA representation.}
    \label{fig:system}
\end{figure}

\subsection{Scaling the Coherent Soft Imitation Learning 
Implementation}
\label{sec:improving}
While \citet{watson2023coherent} demonstrate that CSIL achieves strong empirical results, the implementation struggles to solve more complex tasks in the vision-based setting in RoboMimic.

\textbf{Actor-critic.} To improve on CSIL's base performance, we integrated recent advances from the current state-of-the-art sample-efficient off-policy \ac{RL} methods, which incorporate architectural regularization to facilitate more graceful scaling.
In particular, following the results from XQC \citep{palenicek2025xqc}, we adapt the CSIL critic to a categorical critic and apply batch normalization (BN) \citep{ioffe2015batchnormalizationacceleratingdeep}
and weight normalization (WN) \citep{lyle2024normalization} instead of layer normalization \citep{ba2016layernormalization} to the critic, while applying layer normalization to the policy.
Combining batch normalization and weight normalization improves the optimization landscape and helps maintain a steady `effective' learning rate \citep{lyle2024normalization,palenicek2025CrossQWN}.
We observed that the CSIL implementation by Watson et al., which is based on SAC, suffered from exploding parameter values during training, which impedes learning as the `effective' learning rate drops. 
Incorporating BN and WN into CSIL allowed us to mitigate these numerical issues and improve optimization and scaling.
Incorporating this regularization also enables us to increase the update-to-data (UTD) ratio, where the critic is updated several times per environment step to accelerate learning.
Following \citet{fujimoto2018addressingfunctionapproximationerror}, we use policy delay and only update our policy once per environment step.

\textbf{KL estimation.} Additionally, while \citet{watson2023coherent} use a Monte Carlo KL estimator for the BC regularization term in the actor objective, we identify that this KL can be computed exactly between the latent Gaussian actions due to the shared bijection between policies
\citep{polyanskiy2020infotheory_l1}.

\textbf{Image encoder.} To improve CSIL's performance in image-based settings, we adapted CSIL's original IMPALA-style encoder \citep{espeholt2018impala} by increasing its size, training an encoder per camera instead of one for all cameras, and utilizing spatial softmax pooling \citep{levine2016end} at the end of the network to extract structural information efficiently.

We refer to this improved implementation of CSIL as CSIL++ in the subsequent text.

\begin{figure}
\begin{algorithm}[H]
\KwData{Demonstrations $\gD$, initial temperature $\alpha$, refinement temperature $\beta$,
chunk horizon $H$,\\
base policy $\pi_{\text{LBM}}$,
parametric policy class $\pi_\vphi(\va\mid\vs)$,
prior policy $p(\va\mid\vs)$,
RL steps $N$}
\KwResult{$\pi_{\vphi_N}(\va\mid\vs)$, which, averaged with $\pi_{\text{LBM}}(\va\mid\vs)$, matches or improves the base policy}
Finetune large foundation policy $\pi_{\text{LBM}}$ from expert demonstrations and freeze it\;
Train initial HetStat policy with BC, $\vphi_0 = \argmax_{\vphi} \E_{\vs,\va\sim\gD}[\log \pi_\vphi(\va\mid\vs)]$\;
Define fixed shaped coherent reward $\tilde{r}(\vs,\va) = \alpha\,(\log \pi_{\vphi_0}(\va\mid\vs) - \log p(\va\mid\vs))$\;
Pretrain critic on demonstration data and coherent reward\;
\For{$n \leftarrow 1$ \KwTo $N$}{
    \If{$(n-1) \bmod H = 0$}{
        $\va_{\text{base},n},\dots,\va_{\text{base},n+H-1} \leftarrow \pi_{\text{LBM}}(\vs_n)$\;
    }
    $\va_{\pi,n} \sim \pi_{\vphi_{n-1}}(\cdot\mid\vs_n)$\;
    $\va_{\text{avg},n} \leftarrow \frac{1}{2}(\va_{\text{base},n} + \va_{\pi,n})$\;
    Compute $Q_n$ and $\pi_{\vphi_n}$ using soft off-policy RL, e.g. SAC or XQC, with temperature $\beta$.
}
\caption{CSIL-based finetuning of large behavior models.}
\label{alg:csil}
\end{algorithm}
\vspace{-1.5em}
\end{figure}

\section{Coherent Finetuning of Pretrained Policies}
\label{sec:finetuning}

This section outlines how CSIL is adapted to improve pretrained action-chunking policies, in particular \acp{LBM}.
A high-level description is provided in \Figref{fig:system} and Algorithm \ref{alg:csil}.
An important design decision for LBM finetuning is the choice of optimized action.
This action could be the environment action, a residual correction term to the LBM (\Eqref{eq:residual}), or an averaged action between the base action and the predicted action (\Eqref{eq:average}):

\begin{minipage}{.5\linewidth}
\begin{equation}
    \va_{\text{res}} = \va_{\text{base}} + \va_{\pi}
    \quad
    \text{(residual)},
\label{eq:residual}
\end{equation}
\end{minipage}%
\hfill
\begin{minipage}{.5\linewidth}
\begin{equation}
\va_{\text{avg}} = \textstyle\frac{1}{2}(\va_{\text{base}} + \va_{\pi})
\quad
\text{(average).}\label{eq:average}
\end{equation}
\end{minipage}

\begin{figure}[b]
    \centering
    \includegraphics[width=\linewidth]{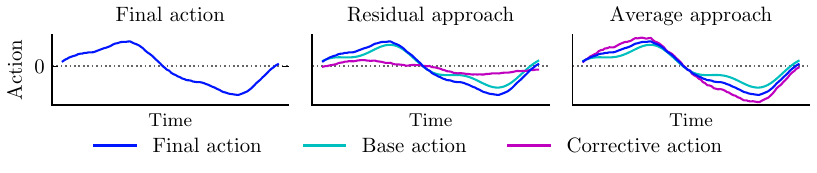}
    \vspace{-2em}
    \caption{Visualization of the different action modalities.
    The cyan, magenta, and blue lines represent the actions generated by the base policy, the predictions of the trainable policy, and the final action executed in the environment, respectively.
    The final action for the residual approach is calculated using \Eqref{eq:residual}, while the total action using the average approach is calculated via \Eqref{eq:average}.
    The average approach improves CSIL's coherent reward by differentiating between expert and non-expert actions more strongly.}
    \label{fig:action-visualization}
\end{figure}

\begin{figure}[!tb]
    \vspace{-3em}
    \centering
    \includegraphics[width=\linewidth]{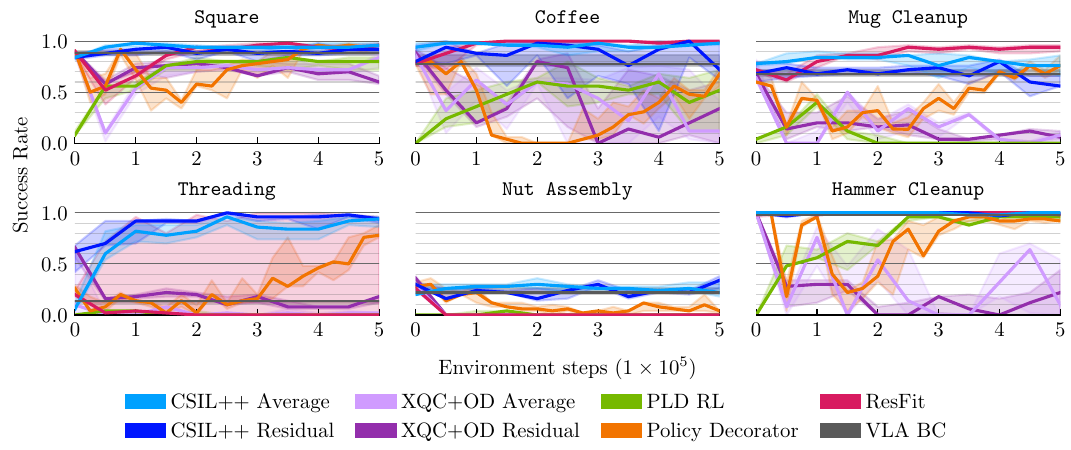}
    \vspace{-2em}
    \caption{Performance of LBM finetuning on six simulated environments across five seeds. Success rates are reported as medians, and uncertainty ranges are quartiles on six sparse reward manipulation tasks and are computed over 50 evaluations every 50k steps.}
    \label{fig:lbm-results}
    \vspace{-1em}
\end{figure}

See \Figref{fig:action-visualization} for a visualization of both.
Directly adapting the base policy to predict environment actions requires optimizing a model with billions of parameters via RL. Even with LoRA \citep{hu2022lora}, this approach is computationally prohibitive due to a massive increase in forward passes of the \ac{LBM}, as the model must be evaluated for every batch sample and environment step.
To learn an additional, smaller corrective policy, we use a frozen version of the \ac{LBM} to predict an action chunk every $H$ steps, while the smaller model is trained to predict a full action every step.
We use \texttt{pi-0.5} in our experiments, where $H{\,=\,10}$.
Using a residual action, i.e., predicting a correction \citep{silver2018residual} on top of the base LBM policy, performed worse (see \Figref{fig:lbm-results}).
For CSIL, we found that a residual BC policy produced a poorly performing coherent reward.
We attribute this result to the residual policy modeling function approximation errors rather than the demonstration behavior, which makes it harder to differentiate expert and non-expert behavior.
We show this experimentally in Figure \ref{fig:reward-hist}.
To mitigate this issue, we take an averaging approach where the RL policy also learns the expert behavior and is averaged with the LBM action.
Moreover, rather than using the LBM's internal representation for RL, we found it better and faster to let the RL policy learn its own observation encoder, which is smaller than the LBM's and therefore easier to finetune. The observation encoder is learned using BC and then kept frozen during RL.

\begin{figure}[!tb]
    \centering
    \vspace{-2em}
    \includegraphics[width=1.0\linewidth]{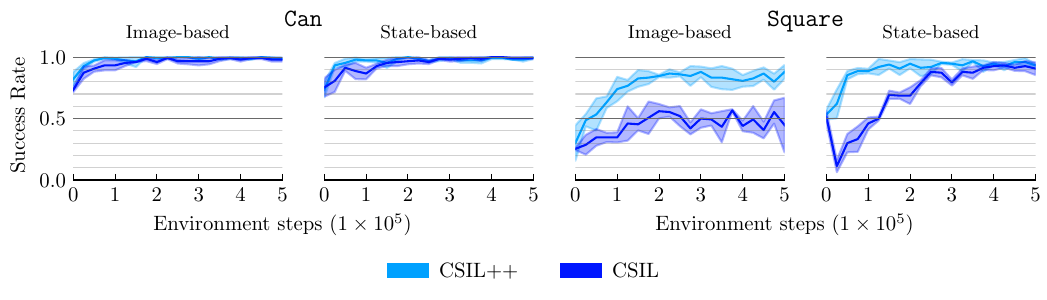}
    \vspace{-2.5em}
    \caption{Performance of CSIL and our improved version CSIL++ on the \texttt{Can} and \texttt{Square} task.
    The `image-based' run takes images and the robot's proprioceptive state as observations.
    The `state-based' run operates on full state information comprising proprioception and ground-truth information of the object.
    In both settings, CSIL++ is more sample-efficient and improves performance significantly in the image-based setting.}
    \label{fig:csil-ablation}
    \vspace{-1em}
\end{figure}

\section{Experimental Setup}

We test our approach on tasks from RoboMimic \citep{robomimic2021} and MimicGen \citep{mandlekar2023mimicgen}, which vary in terms of horizons, precision requirements, and complexity.
Our RL baselines use a constant negative reward of $-1$ as the task feedback for MimicGen environments, and we wish to solve the task as quickly as possible. All our experiments utilize successful termination for bootstrapping.
We train all methods across five independent seeds for 500k steps plus 10k warmup steps and evaluate the policy every 50k steps for 50 episodes.
We use \texttt{pi-0.5} \citep{intelligence2025pi05visionlanguageactionmodelopenworld} as our base model with an action chunk horizon of 10.
 
\textbf{Tasks.}
We use the Robomimic and MimicGen benchmark suite to evaluate our method, due to their breadth in difficulty beyond straightforward pick-and-place tasks.
All tasks use a Franka robot with a parallel-jaw gripper and have an action space dimension of 7: 6 for the end-effector pose delta and 1 for the gripper command.
For \texttt{Square}, we use the released expert demonstration dataset with 200 episodes.
For MimicGen tasks, we use their dataset of 200 episodes generated from a source dataset of 10 demonstrations. 
We use the environment's success flag to  trigger an absorbing state. 

\textbf{Pretraining. } For the simulation experiments, we finetune \texttt{pi-0.5} for 30k steps with a batch size of 128 for 13 hours on 4 H100 GPUs, combining all single-task datasets into one dataset. For the real-world experiments, we train 6 \texttt{pi-0.5} models in increments of 10k expert demonstration samples, starting at 10k samples using the same hyperparameters as for the simulation tasks. We use the model trained on 10k expert demonstration samples for our RL results.

\textbf{Baselines and Ablations.} 
Our primary policy improvement algorithm is CSIL++, as described in Section \ref{sec:improving}.
As prior work uses state-of-the-art, sample-efficient, off-policy RL, we use XQC augmented with the expert offline data (XQC+OD) as an RL baseline and also an ablation of the coherent reward, as we provide XQC with the same HetStat policy, BC pretraining, and initial replay buffer as CSIL++, so only the reward and KL regularization are changed.
We baseline our results with related works ResFit, policy decorator, and PLD. 
We adapt ResFit's and policy decorator's codebases to evaluate them on our chosen six environments and to use \texttt{pi-0.5} as the base model for fair comparison. Since there is no public code available for PLD, we implement the RL component of it in our codebase.
We omit BC probing and VLA distillation to isolate the performance of the RL finetuning, since we believe it to be orthogonal to RL finetuning performance.
Finally, to investigate the effect of the algorithm improvements introduced in Section \ref{sec:improving}, we compare the base version of CSIL with CSIL++ on state- and image-based settings of \texttt{Can} and \texttt{Square} without an LBM.

\textbf{Real-world setup and task.}
Our real-world setup uses a single Franka FR3 arm, a static and a wrist-mounted ZED camera, and a 3D-printed cylinder as its end-effector.
The task is to push a cube over the finish line while staying inside the track markings, requiring reactivity and a high degree of error correction since the block can easily slide off the rounded end-effector. We collect expert demonstrations via teleoperation. The robot is controlled using 2D operational space control at 30 Hz.
RL training is done using a separate data collection process and a learning process, which are run in parallel, similar to SERL \citep{luo2024serl}. Only the entropy coefficient is changed from 0.3 to 0.1 in comparison to the hyperparameters of the simulation experiments.

\textbf{Implementation details and hyperparameters.}
We summarize the specific architectural choices and hyperparameters used across our experiments in Table \ref{tab:hyperparameters} in Appendix \ref{sec:exp-details}.

\section{Experimental Results}
\label{sec:results}

\begin{table*}[b]
\vspace{-1em}
\centering
\resizebox{\linewidth}{!}{%
\begin{tabular}{lcccccc}
\toprule
\textbf{Method} & \texttt{Square} & \texttt{Coffee} & \texttt{Mug Cleanup} & \texttt{Threading} & \texttt{Nut Assembly} & \texttt{Hammer Cleanup} \\ \midrule
VLA & 0.80 [0.70, 0.90] & 1.10 [0.70, 1.20] & 1.60 [1.40, 1.70] & 4.30 [4.30, 4.40] & 3.90 [3.70, 3.90] & 0.10 [0.10, 0.10] \\
VLA (2K demos) & N/A & 0.50 [0.30, 0.57] & 1.20 [1.02, 1.40] & 4.90 [4.83, 4.90] & 2.30 [2.30, 2.40] & 0.00 [0.00, 0.08] \\
\midrule
CSIL++ Avg. & \textbf{0.29 [0.29, 0.32]} & \textbf{0.24 [0.12, 0.24]} & 1.06 [0.75, 1.37] & 1.04 [1.02, 1.12] & \textbf{3.60 [3.53, 3.83]} & 0.03 [0.03, 0.08] \\
CSIL++ Res. & 0.58 [0.44, 0.62] & 0.94 [0.37, 1.56] & 1.56 [1.37, 1.71] & \textbf{0.71 [0.38, 0.75]} & 3.83 [3.77, 3.85] & \textbf{0.04 [0.03, 0.04]} \\
Policy decorator & 1.31 [1.14, 1.34] & 3.27 [3.18, 3.39] & 2.83 [2.72, 2.89] & 3.67 [3.64, 3.75] & 4.50 [4.44, 4.55] & 1.32 [1.13, 1.48] \\
ResFit & 0.64 [0.59, 0.75] & 0.17 [0.11, 0.25] & \textbf{0.73 [0.57, 0.87]} & 4.90 [0.50, 4.93] & 4.93 [4.92, 4.93] & 0.04 [0.04, 0.05] \\
XQC+OD Avg. & 1.79 [1.67, 1.88] & 3.12 [2.55, 3.47] & 4.12 [3.88, 4.16] & 4.83 [4.75, 4.95] & 4.93 [4.92, 4.94] & 3.54 [3.50, 3.62] \\
XQC+OD Res. & 1.46 [1.44, 1.48] & 3.06 [2.12, 3.67] & 4.10 [3.87, 4.46] & 4.16 [4.01, 4.21] & 4.92 [4.91, 4.92] & 3.74 [3.58, 4.09] \\
PLD RL (Res.) & 1.48 [1.39, 1.59] & 2.75 [2.45, 2.85] & 4.63 [4.63, 4.79] & 4.96 [4.94, 4.98] & 4.96 [4.96, 4.98] & 1.19 [1.05, 1.19] \\
\bottomrule
\end{tabular}
}
\caption{LBM finetuning cumulative regret reported as median and quartiles on six sparse-reward manipulation tasks.
The cumulative regret measures sample efficiency using the area between the success rate curve and that of a policy which succeeds in every episode over the first $5 \times 10^5$ environment steps in units of $10^5$ steps, so it ranges from 0 (always successful) to 5 (never successful). Bold indicates the lowest upper quartile. Results aggregate five seeds per environment and method.
}
\label{tab:accum_regret_percentiles}
\end{table*}

To evaluate CSIL++ finetuning of VLAs, our empirical evaluation is guided by four questions:
\begin{enumerate}
    \item How does the performance of CSIL++ compare to CSIL?
    \item Can CSIL++ improve the performance of LBMs better than RL with sparse rewards?
    \item Is RL more sample efficient than BC with additional expert demonstrations?
    \item How do residual and average policies compare across methods?
\end{enumerate}

\textbf{Comparing CSIL and CSIL++.}
While CSIL can be used in the pixel-based setting, we found it struggled to recover state-based performance in harder tasks. 
Figure \ref{fig:csil-ablation} compares CSIL and CSIL++ on RoboMimic's \texttt{Can} and \texttt{Square} tasks.
We can see that while CSIL and CSIL++ closely match on the simpler \texttt{Can} task, there is a much larger gap on the \texttt{Square} task.
CSIL++ is far more sample-efficient in the image-based setting, reflecting the findings of \citet{palenicek2025xqc}. 

\textbf{Finetuning large behavior models.}
We now look at CSIL++ as a means of improving \texttt{pi-0.5} performance, baselining our results using XQC, PLD, policy decorator, and ResFit. \Figref{fig:lbm-results} shows that CSIL++ provides the best improvement in four of the six tasks. Most notably, CSIL++ improves success rates on \texttt{Threading} from 14\% to 92\%. Following CSIL's derivation for coherent improvement of the BC policy, CSIL++ is the only method that matches or improves on the base VLA's performance and does not exhibit initial performance drops across all tasks.
For \texttt{Nut Assembly}, a challenging long-horizon task, while CSIL++ is not able to make significant progress on the task, it is the only method not to unlearn. 
ResFit is the strongest baseline, showing stable improvements and matching CSIL++'s performance on \texttt{Square}, \texttt{Coffee} and \texttt{Hammer Cleanup} while outperforming it on \texttt{Mug Cleanup}. It exhibits initial unlearning in all tasks but \texttt{Coffee} and \texttt{Hammer Cleanup}.
Policy decorator's performance is unstable for all tasks except Threading. In particular, for the saturated task \texttt{Hammer Cleanup}, it exhibits jumps from nearly optimal performance to roughly 10\% within 50k steps. 
The XQC baseline shows that residual policies are useful in the pure RL setting, where residual XQC outperforms average XQC on \texttt{Threading} and \texttt{Square}. The RL component of PLD exhibits immediate unlearning across all tasks, matching the results from \citet{xiao2026_pld}. While it matches other baselines on \texttt{Hammer Cleanup} in its final performance, it requires $\sim$300k environment steps to recover VLA performance. 

\begin{figure}[!t]
    \vspace{-2em}
    \centering
    \begin{minipage}[b]{0.35\linewidth}
        \centering
        \includegraphics[width=0.9\linewidth]{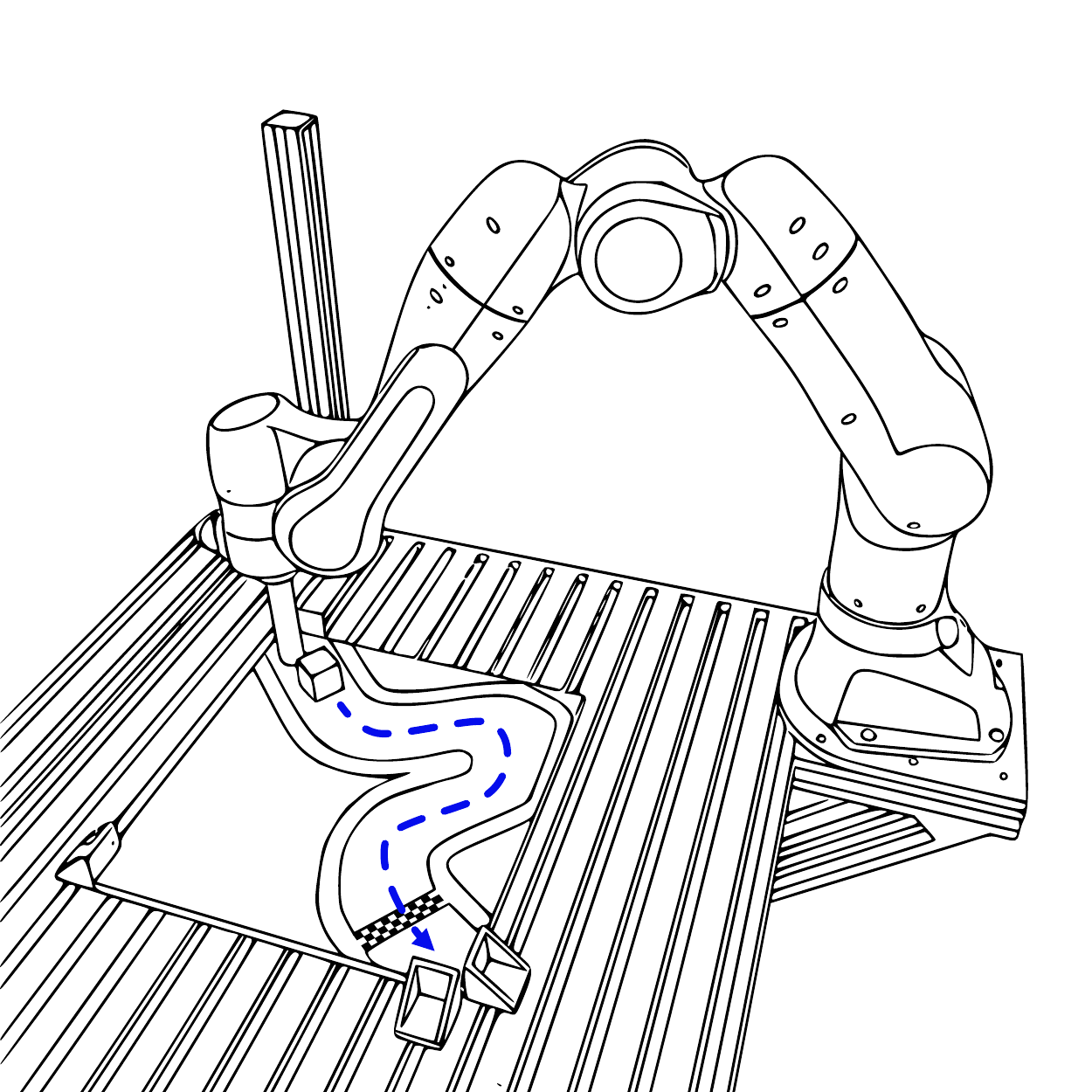}
    \end{minipage}
    \hfill
    \begin{minipage}[b]{0.64\linewidth}
        \centering
        \includegraphics[width=0.8\linewidth]{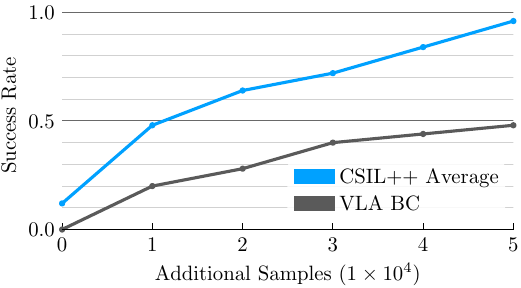}
    \end{minipage}
    \vspace{-1em}
    \caption{
    (left) Illustration of the real-world setup where the robot needs to push the cube over the checkered line without leaving the track markings.
    (right) Evaluation results of the real-world experiments with 25 evaluation episodes every 10k steps. CSIL++ Average achieves near-optimal performance within 50k environment steps.
    In contrast, \texttt{pi-0.5} only achieves roughly 50\% success rate when trained on an additional 50k expert demonstration samples.
    }
    \label{fig:real-world}
    \vspace{-2em}
\end{figure}

In the real-world setting,
\Figref{fig:real-world} shows CSIL++ is able to reach near-optimal performance in the real world within 50k samples, which corresponds to around 28 minutes interaction time.

\textbf{Evaluating demonstration efficiency.}
We collect 2K demonstrations using MimicGen's pipeline for all MimicGen tasks and finetune \texttt{pi-0.5} for 300k steps.
2K is x10 more data than the initial LBM and closely upper-bounds the data collected during RL.
Table \ref{tab:accum_regret_percentiles} shows that the additional demonstrations improve the performance for evaluated tasks except \texttt{Threading}.
However, only on \texttt{Nut Assembly} do the additional demonstrations outperform the RL methods and CSIL++ average in particular.
CSIL++ uses less than 2K total episodes for all tasks except \texttt{Threading} and \texttt{Hammer Cleanup}, where 2K total episodes are reached around 400K environment steps for both.

Comparing the real-world's CSIL++ performance against BC with equal demonstration data, \Figref{fig:real-world} supports the simulation results' findings of superior sample efficiency of RL data collection.
Our finetuned VLA achieves 50\% success rate when given the same amount of additional demonstration samples as CSIL++ uses to solve the task.

\begin{figure}[!t]
    \centering
    \includegraphics[width=1.0\linewidth]{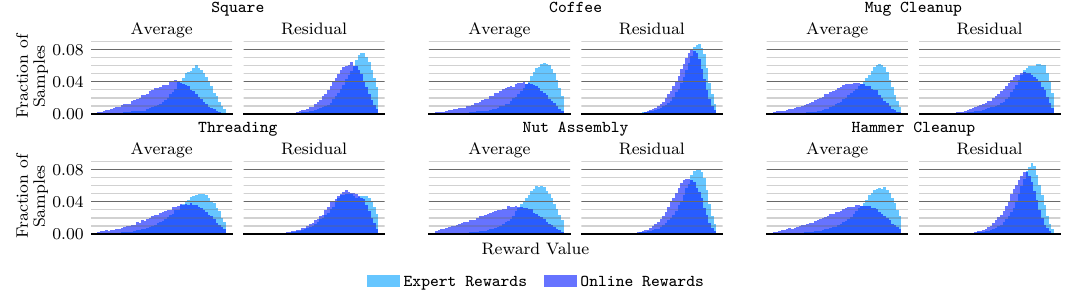}
    \vspace{-2em}
    \caption{
    A comparison of the online and expert reward for the average and residual policy. The `average' approach differentiates between online and expert samples much more strongly than the `residual' approach, which we believe explains its coherent reward performance being better or on par with respect to downstream success rate.
    Online samples are collected per environment using a fixed HetStat policy, trained on expert demonstration data, in combination with a VLA.
    }
    \label{fig:reward-hist}
    \vspace{-1em}
\end{figure}

\textbf{Comparing residual and average policies across methods.} 
To get a clearer view of the benefits and drawbacks of the average approach for finetuning LBMs, we now compare its performance to the residual approach. \Figref{fig:lbm-results} shows CSIL++ Average to be stable across all six environments whereas CSIL++ residual's performance is unstable on \texttt{Coffee} and \texttt{Mug Cleanup}.
In comparison, baselines such as the naive implementation of the average and residual approach as well as policy decorator are much more unstable, especially on \texttt{Hammer Cleanup} and \texttt{Coffee}. We find ResFit to be stable on all tasks except \texttt{Nut Assembly} and \texttt{Threading}.
To understand the superior performance of the average approach over the residual approach on most of the tasks, we plot the distribution of rewards for the expert demonstration samples and online samples per environment and approach collected using policies trained only via BC.
Figure \ref{fig:reward-hist} shows that CSIL++ average differentiates more strongly than CSIL++ residual between online samples and expert samples across the environments despite having equal or higher success rates on them.

\begin{figure}[!b]
    \centering
    \includegraphics[width=1.0\linewidth]{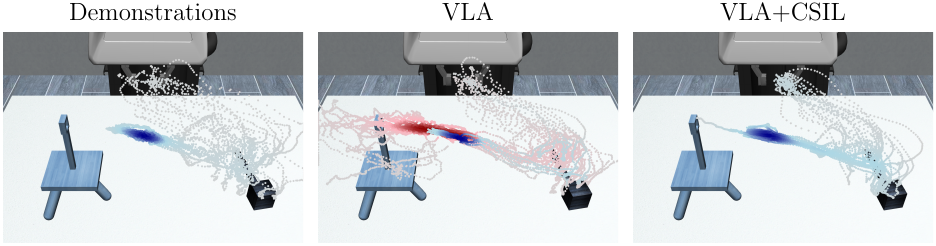}
    \caption{
    A visualization of 25 rollouts between the VLA and the refined policy on \texttt{Threading}, highlighting successful (blue) and failed (red) episodes.
    The base VLA fails the task due to high variance in the insertion trajectory. CSIL++ Average learns to align properly for precise insertion.
    }
    \label{fig:csil-viz}
\end{figure}

\textbf{Visualizing refinement.}
In \Figref{fig:csil-viz}, we visualize the effect of CSIL++ refinement on the VLA behavior, specifically focusing on the \texttt{Threading} task where CSIL++ had the largest impact. 
We see that \texttt{pi-0.5}'s main failure mode is high trajectory variance around insertion. 
CSIL++ refinement learns the alignment behavior pre-insertion that is required for consistent success.

\section{Conclusion}

We demonstrate that inverse reinforcement learning can be used to finetune large behavior models, as CSIL provides a coherent, dense reward that mitigates the sample inefficiency and instability typically associated with refining \acp{LBM} using RL with sparse rewards.
By replacing sparse environment signals with a dense reward derived from expert demonstrations, we are able to achieve policy improvement without expensive unlearning.
We have also shown that CSIL can be scaled to harder vision-based tasks by incorporating innovations in deep actor-critic methods.
Our experimental study pushes LBM finetuning beyond pick-and-place tasks and demonstrates that CSIL-based finetuning can provide dramatic improvements in learning precise insertion tasks such as the \texttt{Threading} task, and also scales to real-world deployment.
For future work, our results show there is still further research needed to achieve greater learning stability and sample-efficiency in LBM finetuning, in particular for long-horizon tasks as seen in the \texttt{Nut Assembly} task.

\textbf{Limitations.} A fundamental limitation of CSIL is how it leverages BC.
If there are few demonstrations such that the BC policy is weak, CSIL's sample efficiency will also weaken.
CSIL also optimizes an imitation objective, so it is susceptible to suboptimal demonstrations, as it will reinforce any suboptimal behaviors, e.g., hesitation, potentially impacting performance.


%

\subsection*{AI use statement}

In this work, we used generative AI tools for aiding with software development and   providing feedback on research  methodology, experiment design and this paper.
We have not used generative AI tools for synthetic datasets, data processing, or data interpretation.
Commenting on aid for novel mathematical results is not relevant to this work.
Additionally, we used generative AI tools for help with creating the Figure \ref{fig:real-world} illustration from a photo reference of the experimental setup.
We have reviewed all AI-assisted work, primarily through inspection by the lead authors.
We take responsibility for the final content of this work,
including text, claims or artifacts produced with the aid of generative AI.

\subsection*{Reproducibility statement}

We use open-source simulation benchmarks with freely available demonstration data sets for reproducibility.
Our code is anonymized and provided as supplementary material for reference.
Our real-world experiment and demonstration data is not possible to open-source due to a bespoke lab setup, including a custom built table and camera rig, but is described as faithfully as possible.

\subsection*{Acknowledgments}
We wish to thank Florian Vogt for his assistance implementing XQC.
Christian Scherer is supported by the Konrad Zuse School of Excellence in Learning and Intelligent Systems (ELIZA) through the DAAD programme Konrad Zuse Schools of Excellence in Artificial Intelligence, sponsored by the Federal Ministry of Education and Research. This research was funded by the research cluster “Third Wave of AI”, funded by the excellence program of the Hessian Ministry of Higher Education, Science, Research and the Arts, hessian.AI and by the Deutsche Forschungsgemeinschaft (DFG, German Research Foundation) under Germany’s Excellence Strategy (EXC-3057/1 “Reasonable Artificial Intelligence”, Project No. 533677015). It was further supported by a UKRI/EPSRC Programme Grant [EP/V000748/1] and partially supported by the German Federal Ministry of Research, Technology and Space (BMFTR) under the Robotics Institute Germany (RIG). Ingmar Posner holds concurrent appointments as a Professor
of Applied AI at the University of Oxford and as an Amazon Scholar. This paper describes work performed at the University of Oxford and is not associated with Amazon.


\bibliography{lib.bib}
\bibliographystyle{iclr2027_conference}

\appendix

\section{Coherent Soft Imitation Learning}

For MDPs with success-based absorbing states, like many manipulation tasks, IRL methods can struggle if their rewards are positive.
Agents are incentivized to `stay alive' and accrue reward, and therefore never complete the task and enter the absorbing state.

To prevent this behavior in CSIL, the coherent reward can be bounded and its value subsequently shifted by its maximum reward, to ensure the coherent reward is always negative.
To achieve this bounding of the maximum reward, the authors add a small bias term $\sigma_{\min}^2$ to the predictive variance that enables them to define an upper bound for the likelihood. As the $\tanh$ transformation is also bounded in practice to ensure numerical stability, the authors are then able to derive the following upper bound for the reward $r(\rvs,\rva)$, given action dimension $d_a$, a uniform prior across the action range $[-1, 1]^{d_{a}}$, and $\alpha=d_a^{-1}$ 
\begin{align*}
    r(\rvs,\rva) \leq -0.5 \cdot\log\left(\pi\sigma^2_{min}/2\right) + \tilde{c},
\end{align*}
where $\tilde{c}$ depends on the $\tanh$ clipping term \citep{watson2023coherent}.

\section{Experimental Details}
\label{sec:exp-details}

\subsection{Methods, Baselines and Ablations}

\textbf{CSIL.}
We follow the default hyperparameters and architectural choices described in \citet{watson2023coherent}, changing only the total number of environment steps to 510k. 

\textbf{CSIL++.} We use the hyperparameters described in Table \ref{tab:hyperparameters}. In particular, we do not uniformly distribute the bins of the categorical critic. Instead, we calculate the Monte Carlo values for the demonstrations and then distribute the bins using quantiles. We normalize the reward and the entropy coefficient by the standard deviation of the Monte Carlo values. The residual is not conditioned on the base policy's actions, while the critic uses the current state as well as the combined action of the base policy and residual policy as input. We replace the upper bound of the coherent reward function with the maximum reward computed over the demonstration data and additionally clip the reward to ensure all reward values are negative. 

\textbf{XQC + OD.} We use the same network sizes, learning rates, and pretraining hyperparameters as for CSIL++. We distribute the bins of the categorical critic uniformly. Since we pretrain the policy, we are able to initialize the replay buffer with on-policy actions without a loss of diversity in data. The residual is not conditioned on the base policy's actions, while the critic uses the current state as well as the combined action of the base policy and residual policy as input.

\textbf{PLD.} We closely follow the hyperparameters from \citet{xiao2026_pld}.
Our adjustments adapt the scaling of the residual actions to the range $[-0.1, 0.1]$ and use an initial entropy coefficient of 0.1. 

\textbf{ResFit.} We use the public code base\footnote{\url{https://github.com/amazon-far/residual-offpolicy-rl}} of \citet{ankile2025residual}, changing only the environment wrappers to run the experiments on RoboMimic and MimicGen environments and the base policy implementation to be able to use \texttt{pi-0.5} as a base model for fair comparison.

\textbf{Policy decorator.} Similar to ResFit, we adapt the public code base\footnote{\url{https://github.com/tongzhoumu/policy_decorator}} of \citet{yuan2025decorator} to run the experiments on RoboMimic and MimicGen environments and use \texttt{pi-0.5} as a base model for fair comparison. We adapted the initial entropy coefficient from $e^1$ to $e^{0.2}$ as we observed large initial performance losses due to a too strong entropy regularization.

\newpage 

\subsection{Hyperparameters}

\begin{table}[h]
\centering
\caption{Core Hyperparameters for CSIL++}
\begin{tabular}{lc}
\midrule
\textbf{Hyperparameter} & \textbf{Value} \\
\midrule
\multicolumn{2}{c}{\textit{Architecture \& Optimization}} \\
\midrule
Batch Size & 128 \\
Critic Network & [1024, 1024]\\
HetStat Network & [1024, 1024, 48, 1024]\\
Learning Rates & 3e-4 \\ 
ResNet Channels & $[16, 32, 64, 128]$ \\
Visual Pooling & Spatial Softmax \\
Critic Normalization & Batch Normalization \\
Policy Normalization & Layer Normalization \\
Categorical Critic Bins & 101 \\
Categorical Critic Range & $[-8, 0]$ \\
Distribution of Bins & By Quantile \\
\midrule
\multicolumn{2}{c}{\textit{Training Dynamics}} \\
\midrule
Warmup Steps & 10,000 \\ 
Total Environment Steps & 510,000 \\ 
Update-To-Data (UTD) Ratio & 4 \\
Policy Delay & 4 \\
Entropy Coefficient & $0.3/\textrm{Std. of MC values}$ \\
Replay Buffer Size & 250,000 \\
\midrule
\multicolumn{2}{c}{\textit{Pretraining}} \\ \midrule
Demonstrations per Env. & 200 \\
Policy Pretraining Steps (Image) & 25,000 \\
Policy Pretraining Steps (State) & 50,000 \\ 
Critic Pretraining Steps & 20,000 \\
Environment Horizon & Standard (Task-dependent) \\
\midrule
\end{tabular}
\label{tab:hyperparameters}
\end{table}

\newpage

\section{Additional Experimental Results}

\begin{table*}[h]
\centering
\resizebox{\linewidth}{!}{%
\begin{tabular}{lcccccc}
\toprule
\textbf{Method} & \texttt{Square} & \texttt{Coffee} & \texttt{Mug Cleanup} & \texttt{Threading} & \texttt{Nut Assembly} & \texttt{Hammer Cleanup} \\ \midrule
VLA & 0.84 [0.82, 0.86] & 0.78 [0.76, 0.86] & 0.68 [0.66, 0.72] & 0.14 [0.12, 0.14] & 0.22 [0.22, 0.26] & 0.98 [0.98, 0.98] \\
VLA (2K demos) & N/A & 0.90 [0.89, 0.94] & 0.76 [0.72, 0.80] & 0.02 [0.02, 0.04] & 0.54 [0.52, 0.54] & 1.00 [0.98, 1.00] \\
\midrule
CSIL++ Avg. & \textbf{0.96 [0.92, 0.98]} & 0.98 [0.96, 1.00] & 0.80 [0.68, 0.84] & 0.92 [0.90, 0.94] & \textbf{0.27 [0.23, 0.32]} & \textbf{1.00 [0.98, 1.00]} \\
CSIL++ Res. & 0.94 [0.90, 0.96] & 0.78 [0.62, 0.94] & 0.66 [0.52, 0.72] & \textbf{0.94 [0.92, 0.96]} & 0.26 [0.20, 0.30] & \textbf{1.00 [1.00, 1.00]} \\
Policy decorator & \textbf{0.96 [0.96, 0.96]} & 0.68 [0.56, 0.74] & 0.76 [0.72, 0.86] & 0.78 [0.76, 0.88] & 0.04 [0.02, 0.04] & 0.92 [0.92, 0.98] \\
ResFit & \textbf{0.96 [0.96, 0.98]} & \textbf{1.00 [0.98, 1.00]} & \textbf{0.94 [0.90, 0.96]} & 0.00 [0.00, 0.96] & 0.00 [0.00, 0.00] & 0.98 [0.98, 1.00] \\
XQC+OD Avg. & 0.82 [0.74, 0.88] & 0.05 [0.00, 0.43] & 0.10 [0.08, 0.12] & 0.00 [0.00, 0.02] & 0.00 [0.00, 0.00] & 0.12 [0.00, 0.56] \\
XQC+OD Res. & 0.66 [0.58, 0.74] & 0.28 [0.00, 0.60] & 0.12 [0.08, 0.18] & 0.12 [0.10, 0.20] & 0.00 [0.00, 0.00] & 0.20 [0.00, 0.42] \\
PLD RL (Res.) & 0.82 [0.78, 0.84] & 0.56 [0.26, 0.62] & 0.00 [0.00, 0.00] & 0.00 [0.00, 0.00] & 0.00 [0.00, 0.00] & 0.96 [0.94, 0.98] \\
\bottomrule
\end{tabular}
}
\caption{LBM finetuning success rates reported as median and quartiles on six sparse-reward manipulation tasks.
Bold indicates highest median seen over training. Success rates are computed from 5 evaluations with 50 episodes each, and aggregate results across five seeds per environment and method are concatenated.
}
\label{tab:success_rates_percentiles}
\end{table*}

\begin{figure}[h]

    \centering
    \includegraphics[width=\linewidth]{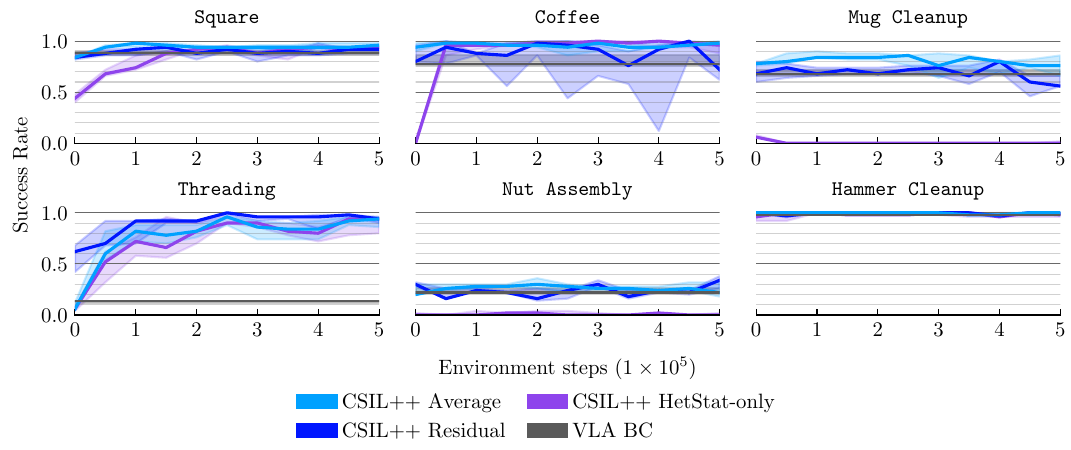}
    \vspace{-2em}
    \caption{Comparison of CSIL++'s performance on LBM finetuning and HetStat-only finetuning across five seeds. Success rates are reported as medians, and uncertainty ranges are quartiles on six sparse reward manipulation tasks and are computed over 50 evaluations every 50k steps.}
    \label{fig:lbm-results-hetstat-only}
\end{figure}

\end{document}